# L-LO: Enhancing Pose Estimation Precision via a Landmark-Based LiDAR Odometry

Feiya Li, Chunyun Fu, and Dongye Sun

*Abstract* — The majority of existing LiDAR odometry solutions are based on simple geometric features such as points, lines or planes which cannot fully reflect the characteristics of surrounding environments. In this study, we propose a novel LiDAR odometry which effectively utilizes the overall exterior characteristics of environmental landmarks. The vehicle pose estimation is accomplished by means of two sequential pose estimation stages – horizontal pose estimation and vertical pose estimation. To achieve effective landmark registration, a comprehensive index is proposed to evaluate the level of similarity between landmarks. This index takes into account two crucial aspects of landmarks – dimension and shape – in evaluating their similarity. To assess the performance of the proposed algorithm, we utilize the widely recognized KITTI dataset as well as experimental data collected by an unmanned ground vehicle platform. Both graphical and numerical results indicate that our algorithm outperforms leading LiDAR odometry solutions in terms of positioning accuracy.

## I. Introduction

SIMULTANEOUS localization and mapping (SLAM) is pivotal for various applications such as robot navigation, autonomous driving, and 3D map reconstruction. The existing SLAM algorithms in the literature can be generally classified into three major categories, including LiDAR-SLAM (L-SLAM) [1] [2], visual-SLAM (V-SLAM) [3] and RGB-D visual SLAM [4]. Recently, L-SLAM has attracted increasing attention due to the fast commercialization of LiDAR sensors and their valuable advantages such as high range accuracy [5], large field of view (FOV) [6], and high robustness against illumination and weather changes[7]. Currently, most existing L-SLAM solutions are based on geometric feature registration or occupancy grid registration, both of which can be utilized to achieve localization and mapping in robotic and autonomous driving applications. However, the former methods are only able to generate maps with sparse geometric features or maps that contain partial information of the surrounding environment. On the other hand, the latter approaches provide occupancy maps that can only be used to determine the existence of obstacles at different locations, and these maps fail to provide sufficient details of the environment. Moreover, the existing L-SLAM algorithms still suffer from insufficient localization accuracy in complex scenarios.

Aiming to address the above limitations in the existing approaches, we introduce in this paper a novel LiDAR odometry framework. This method not only provides a heightened level of localization precision based on landmark registration, but also constructs a richly detailed global map composed of extracted landmarks. Our proposed paradigm consists of three major components: data pre-processing, LiDAR odometry (LO), and mapping. Each of these components can be further divided into several sub-components, as shown in Fig. 1.

In the data pre-processing stage, we employ the segmentation algorithm introduced in [8] to differentiate between ground and non-ground points. For the non-ground points, clustering is executed based on Euclidean distances, as explained in [9]. After clustering is completed, outliers are filtered using the statistical outlier removal method introduced in [10]. Upon cleansing outliers, less noisy and more reliable clusters can be achieved. Then, initial landmark matching is accomplished by measuring the spatial distances between cluster centers across consecutive frames.

In the LO stage, for each pair of landmarks successfully paired in the above "initial landmark matching" step, we partition the matched clusters into two layers in the vertical direction – the upper layer and the lower layer. Then, for each layer, we construct convex hulls for the 3D points contained in this layer, in order to reflect the exterior profiles of the paired landmarks. On this basis, a comprehensive similarity index is designed to evaluate the level of similarity between matched convex hulls, and then a cost function is established for the more similar pair of convex hulls, with the vehicle horizontal pose change being the argument of this cost function. This cost function is then optimized using the coordinate rotation method [11] to achieve the optimal horizontal pose estimation. Afterwards, the vehicle pose in the vertical plane, including pitch angle and vertical position, is estimated sequentially using the ground points and the horizontal pose estimation results.

In the mapping stage, using the above-obtained 6-DOF vehicle pose, landmark points and ground points, we transform the 3D points in each frame into a unified world coordinate system via transform integration [12]. Subsequently, voxelization is performed to achieve a final point cloud map containing detailed environment information in a lightweight fashion.

This paragraph of the first footnote will contain the date on which you submitted your paper for review, which is populated by IEEE. This work was supported by the Chongqing Technology Innovation and Application Development Project under Grant CSTB2022TIAD-DEX0013. *(Corresponding author: Chunyun Fu.)*

Feiya Li, Chunyun Fu, and Dongye Sun are with the College of Mechanical and Vehicle Engineering, Chongqing University, Chongqing 400044, China (e-mail: FairLee@cqu.edu.cn; fuchunyun@cqu.edu.cn; dysun@cqu.edu.cn).

Color versions of one or more of the figures in this article are available online at http://ieeexplore.ieee.org

The major contributions of this work include: 1) proposing a novel landmark-based LiDAR odometry framework, which comprises two sequential pose estimation modules – a horizontal pose estimation module and a vertical pose estimation module. To the best of our knowledge, this work constitutes the first LiDAR odometry which works based on the overall exterior characteristics of environmental landmarks, rather than simple geometric features in point clouds such as points, lines, and planes. 2) introducing a comprehensive similarity index based on two key factors – landmark size and landmark shape – to evaluate the level of similarity between any two landmarks. 3) designing an adaptive step size that can be dynamically adjusted at each iteration of optimization, which not only facilitates convergence but also reduces computation load.

The proposed algorithm is evaluated based on the commonly used KITTI dataset [13] renowned for its varied and demanding scenarios — and experimental data collected by an unmanned ground vehicle (UGV) platform. The proposed method provides superior positioning accuracy compared with leading SLAM methods in the literature, such as LOAM-Velodyne [12]. Furthermore, our algorithm concurrently constructs a richly detailed map, featuring segmented objects, thereby enabling more comprehensive reflection and interpretation of the real environment.

## II. RELATED WORK

In the existing literature, most L-SLAM solutions are based on geometric feature registration or occupancy grid registration. The former approaches can be further categorized into three main types, including point feature registration [14], line feature registration [15], and plane feature registration [16].

Zhang and Singh [12] proposed a LiDAR odometry and mapping (LOAM) method for real-time applications. This method categorized feature points extracted from LiDAR point clouds into two major types – edge points and planar points – based on the local surface smoothness. These feature points were employed in the LiDAR odometry to estimate the pose of the robot. Based on LOAM, Shan and Englot [17] proposed an improved SLAM method called 'LeGO-LOAM'. In this algorithm, point cloud was firstly clustered to filter out noise, after which the feature points were extracted and evenly distributed in different regions. Then, edge and planar points were used to construct the cost function which was solved by means of the Levenberg-Marquardt (L-M) optimization method to optimize the robot's pose. In unstructured environments, the extracted feature points can become too sparse to ensure effective point feature registration. To tackle this issue, Guo et al. [18] proposed a method to extend the local structure information, in which the local point cloud in the voxels around the feature points and the points with large intensity changes were employed as additional feature points for point feature registration. When dealing with sparse point clouds or environments with large noise and/or clutters, point feature extraction is more prone to errors and abnormalities, which in turn results in more failures of point feature registration. To overcome this limitation, Zhou et al. [19] conducted point cloud segmentation and filtered out unstable objects in the pre-processing stage, and then extracted four types of features including edge features, sphere features, plane features and ground features. With these various features, the LiDAR pose can be accurately estimated by using the truncated least squares method for feature processing.

Apart from point features, line features are also often used as landmarks for registration in scenarios with obvious edges, such as corridors and underground car parks. To deal with underground environments, Wu et al. [20] proposed a SLAM method by combining line features and point features. In this approach, line features were extracted from point cloud by means of segmentation and model fitting, while point features were extracted by using a curvature detector. Then, these two types of features were registered through the commonly used iterative closest point (ICP) algorithm. Park et al. [21] proposed a mobile robot SLAM method based on least-square matching. The line features extracted from point cloud were matched with a map database to realize position estimation of the robot, and the line features were registered and incrementally saved to supplement the map database. Xu et al. [22] proposed a similar approach to incremental map construction, differently, in this work high-dimensional information (e.g. length and slope) of line features were taken into account and an improved high-dimensional likelihood field model was proposed for the pose estimation process.

In the existing L-SLAM literature, apart from point features and line features, plane features have also been used to estimate poses of robots/vehicles, especially in structured environments. Grant et al. [23] proposed a plane-feature-based L-SLAM method, which performed frame-to-frame registration by matching closely located planes that were almost parallel to each other, thereby achieving accurate pose estimation in real-time. To tackle the mapping problem in man-made structured environments, Ćwian et al. [24] proposed the Plane-LOAM method which employed a set of 3D plane features and line features for map representation, and this approach achieved better registration results than LOAM for high-speed vehicle maneuvers. Zhang et al. [25] first conducted point cloud pre-processing and then extracted line features and plane features from the pre-processed points. Besides, time constraints and distance constraints were added in key-frame based loop-closure detection to reduce pose estimation error. Huo et al. [26] obtained line features and plane features by means of segmentation and clustering, and these features were then employed in the L-M optimization method to solve the 6-DOF pose estimation problem. Zhou et al. [27] utilized planes as landmarks in their L-SLAM approach for indoor applications. Global registration was conducted in the front-end of their SLAM algorithm, which achieved plane identification and pose estimation in real-time. In the back-end, plane adjustment was introduced to optimize keyframe poses and plane parameters. Zhou et al. [28] also employed plane features as landmarks to construct an indoor L-SLAM system, which consists of three major parts:

localization, local mapping, and global mapping. The first part conducts global registration for pose estimation, the second part performs optimization of keyframe poses and plane parameters, and the third part further refines keyframe poses and plane parameters. It should be noted that SLAM methods based on geometric feature registration are only able to generate maps with sparse geometric features or maps that contain partial information of the surrounding environment.

Apart from geometric features, occupancy grids [29] are also commonly used to construct maps and estimate vehicle poses in SLAM applications. Occupancy grid maps have been often applied in SLAM methods based on Rao-Blackwellized particle filters (RBPFs). In RBPF-based SLAM solutions, the large number of particles usually leads to heavy computation load as each particle stores the information of an individual map. To address this problem, Grisetti et al. [30] proposed a 'Grid Mapping' method to reduce the number of particles in an adaptive fashion, and designed a precise proposal distribution which considers not only the robot motion but also the latest measurement. This method greatly reduced the uncertainty of the filter in terms of robot pose prediction. Su et al. [31] verified the principles proposed in [30] through both simulation studies and experiments. Hampton et al. [32] proposed an occupancy-grid-based SLAM algorithm under the random finite set (RFS) framework. By modeling sensor measurements as RFSs, this SLAM algorithm achieved high robustness under conditions of false and missed detections. Sugiura and Matsutani [33] proposed an FPGA implementation of L-SLAM based on occupancy grid map. An RBPF was used in this approach to realize registration of multiple particles in parallel, which leads to reduced computation load and improved throughput. It can be understood from the above discussions that SLAM approaches based on occupancy grid registration provide occupancy maps that can only be used to determine the existence of obstacles at different locations of the environment, and these maps fail to provide sufficient details of the environment.

III. METHODOLOGY

In this section, the proposed method is described in details. Our proposed paradigm consists of three major components: data pre-processing, LiDAR odometry (LO), and mapping. Each of these components can be further divided into various detailed stages. Fig. 1 graphically demonstrates the overall architecture of the proposed LO system, offering an in-depth illustration of its contained modules and working flow.

*A. Per-processing*

The data pre-processing component consists of four steps: ground segmentation, clustering, outlier removal, and initial landmark matching. The first step extracts ground points from the entire point cloud, the second step clusters the remaining non-ground points based on their Euclidean distances, the third step filters out noise in each cluster, and the last step performs initial association of clusters in two consecutive frames.

*1) Ground Segmentation*

In the raw point cloud data obtained by a LiDAR sensor, a large portion of points result from the ground in the surrounding environment. To facilitate subsequent classification of objects, the raw point cloud data needs to be segmented into two main categories: ground points and non-ground points.

In this paper, the segmentation algorithm proposed in [8] is used to distinguish these two types of points. With this algorithm, the points corresponding to the ground are extracted through comparisons with local line fits. It should be mentioned that this segmentation method provides impressive real-time performance, which makes it highly suitable for our SLAM application. Interested readers are referred to [8] for more details.

*2) Clustering*

Clustering is a necessary process in L-SLAM applications to classify points based on their properties. Points that share common characteristics are clustered into the same group, which provides useful information regarding object location and its exterior profile.

Note that in real traffic scenarios, the relative distances between major objects (such as buildings, vehicles, and vegetation) must be greater than zero. Hence, in this study, Euclidean Clustering is used to categorize the obtained 3D point cloud into different clusters.

Euclidean clustering is a clustering method based on evaluating Euclidean distances between points, specifically, the points with Euclidean distances less than a certain threshold are classified into the same cluster [9]. By this means, a rough segmentation of the original 3D point cloud is achieved.

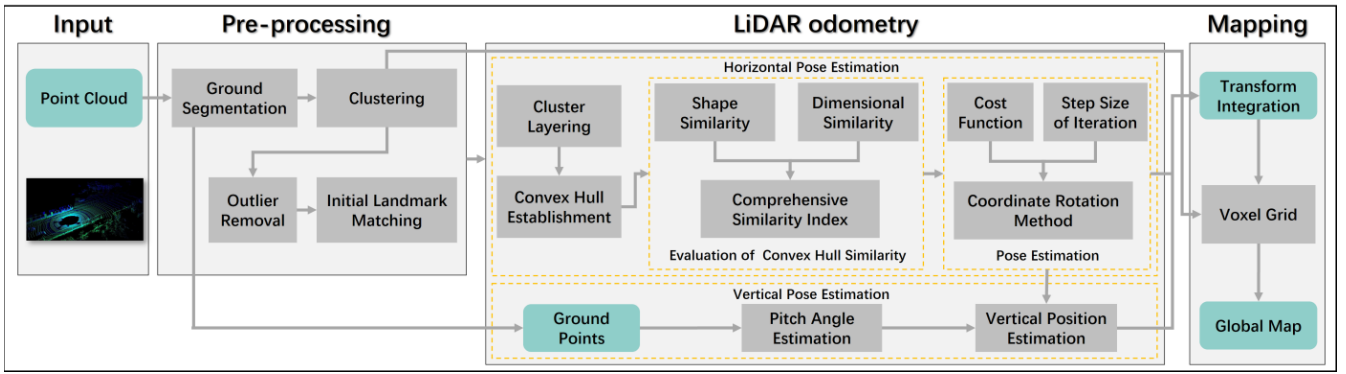

**Fig. 1**. Proposed LiDAR-based odometry systems framework.

*3) Outlier Removal*

The point cloud obtained by a LiDAR sensor is an uneven sampling of the environment being explored. Namely, the objects located closer to the LiDAR sensor produce denser sample points, while those situated farther away give rise to sparser sample points. The denser the point cloud is, the more information it provides. As a result, a sparse point cloud, such as the outliers resulting from sensor noise, provides negligible or even misleading information.

In Euclidean clustering, a fixed threshold is normally employed to search for points around the seed points to achieve clustering, which can lead to outliers in the resulting clusters. To ensure SLAM performance, in this work, we use the statistical outlier removal method proposed in [10] to suppress noise for each cluster of points. It is assumed in this approach that the distances from a certain point to its neighboring points in the same cluster are normally distributed. Then, the average distance from a specific point $p$ to its $K$-neighboring points in the cluster, $d_{avg}$, is defined by:

$$d_{avg} = \frac{1}{K}\sum_{i=1}^{K} d(p, p_i) \quad (1)$$

where $d(p, p_i)$ denotes the distance between point $p$ and its $i$-th neighbor $p_i$ and $K$ represents the total number of neighboring points.

Subsequently, to determine the level of variation in these distances, we calculate their standard deviation $d_{std}$ as follows:

$$d_{std} = \sqrt{\frac{1}{K}\sum_{i=1}^{K}(d(p, p_i) - d_{avg})^2} \quad (2)$$

If a given distance $d(p, p_i)$ exceeds the threshold (i.e. $d_{avg} + 2 \times d_{std}$ in this study), the point $p_i$ is flagged as an outlier and consequently removed from the cluster. By employing this method, reduction of outliers existing in the clustered point cloud is achieved.

*4) Initial Landmark Matching*

Similar to typical L-SLAM algorithms in the literature, our proposed approach is also based on feature registration. Hence, before the actual LiDAR odometry takes place, in this study an initial landmark matching step is executed to establish correspondence between landmarks appearing in two consecutive frames, $k$–1 and $k$.

Firstly, the geometric center of each cluster, which is also called cluster center in this study, is calculated for each cluster. The geometric center of the $r$-th cluster $c_i$ is defined as the average position of all points in this cluster, which is given by:

$$c_r = \left(\frac{\sum_{i=1}^{N} x_i}{N}, \frac{\sum_{i=1}^{N} y_i}{N}, \frac{\sum_{i=1}^{N} z_i}{N}\right) \quad (3)$$

where $(x_i, y_i, z_i)$ represent the coordinates of the $i$-th point in the cluster, and $N$ is the total number of points in the cluster.

After obtaining the cluster centers, we aim to match these centers across two consecutive frames, $k$–1 and $k$. To do this, we evaluate the distance between each pair of cluster centers in frame $k$–1 and frame $k$, which can be expressed as follows:

$$d_{r2s} = \left\| c_r^{k-1} - c_s^k \right\| \quad (4)$$

where $c_r^{k-1}$ represents the center of the $r$-th cluster in frame $k$–1, $c_s^k$ denotes the center of the $s$-th cluster in frame $k$, and $d_{r2s}$ is the Euclidean distance between these two centers.

For a given cluster center $c_r^{k-1}$ in frame $k$–1, we compare all $d_{r2s}, s \in Num^k$ (where $Num^k$ denotes the total number of clusters in frame $k$) and find the shortest distance $d_{r2s\_min}$ among all $d_{r2s}, s \in Num^k$. If $d_{r2s\_min}$ is less than the average of all such computed minimum distances, the two clusters (one from frame $k$–1 and the other from frame $k$) are deemed to be successfully matched [34]. By this means, we can effectively establish correspondences between clusters in consecutive frames based on their geometric centers.

*B. LiDAR Odometry*

In our approach, the odometry is decomposed into two sequential components: horizontal pose estimation module and vertical pose estimation module. Within the 6-DOF of the vehicle motion, we hypothesize a null roll angle for the vehicle to simplify our computation. Note that this hypothesis is sensible as roll angle is usually a crucial variable in vehicle handling dynamics, as opposed to in vehicle localization. The former module of our LO estimates the vehicle pose in the horizontal plane, i.e. the vehicle position in the $x$- and $y$-direction as well as its yaw (heading) angle with respect to its initial pose. The latter module provides the vehicle pose in the vertical plane, namely the vehicle position in the $z$-direction and its pitch angle with respect to its initial pose. These two modules of our proposed LO are explained in detail in the following sections.

*1) Horizontal Pose Estimation*

In this section, we elaborate on the principles of the horizontal pose estimation module in our proposed LO. This module is composed of four steps: cluster layering, convex hull establishment, evaluation of convex hull similarity, and pose estimation. In the first step, each cluster of points is divided to two layers (i.e. an upper layer and a lower layer), according to the average height of points in this cluster. In the second step, the two layers of points are projected onto the $x$-$o$-$y$ plane, and for each layer of projected points, a convex hull encircling all projected points is constructed. In the third step, the similarity between convex hulls formed in two consecutive frames is evaluated in terms of a comprehensive similarity index. In the final step, the coordinate rotation transformation algorithm is employed to achieve pose estimation by maximizing the extent of overlap between similar convex hulls constructed in two consecutive frames.

*a) Cluster Layering*

LiDAR point cloud data portrays the exterior profiles of detected objects. In practice, we often encounter objects whose profile differs in the vertical direction. Taking a parked sedan as an example, its lower body is normally bigger than its upper body. As a result, when using these objects for registration, the profile difference in the vertical direction should be properly dealt with, which otherwise can result in significant registration error and SLAM performance deterioration.

To tackle the above challenge, in this paper, we take into consideration of the object profile difference in the vertical direction, by partitioning each cluster of points into two layers – an upper layer and a lower layer, as shown in Fig. 2.

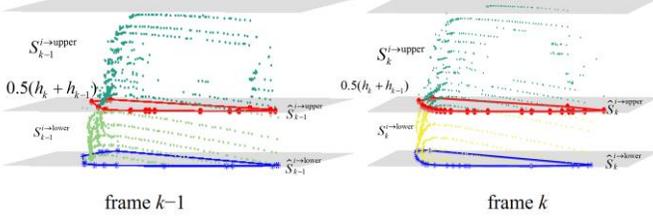

**Fig. 2**. Illustration of Cluster Layering. In frames $k$–1 and $k$, the point cloud data for each cluster is partitioned into two layers – the upper layer and the lower layer. The red closed curve represents the convex hull computed from the point cloud in the upper layer, while the blue closed curve is derived from the point cloud in the lower layer.

The height at which a cluster in frame $k$ should be partitioned is determined by not only itself, but also its matched cluster in frame $k$–1. Specifically, we find the average height of a cluster of points in frame $k$ (denoted by $h_k$), as well as the average height of its matched cluster in frame $k$–1 (denoted by $h_{k-1}$). Then, these two matched clusters are both partitioned at the same height of $0.5(h_k + h_{k-1})$.

By this means, the matched pair of clusters in frames $k$–1 and $k$ are partitioned into four parts: the upper and lower layers of the cluster in frame $k$–1, and the upper and lower layers of the matched cluster in frame $k$. These four layers are used in the following section to construct convex hulls for registration purposes.

*b) Convex Hull Establishment*

Based on the layers we obtained in the previous section, to proceed with registration, we first establish convex hulls for the matched clusters in frames $k$–1 and $k$. For the $i$-th pair of matched clusters in frames $k$–1 and $k$, we denote the upper and lower layers of the cluster in frame $k$–1 by $S_{k-1}^{i \to \text{upper}}$ and $S_{k-1}^{i \to \text{lower}}$, and the upper and lower layers of the cluster in frame $k$ by $S_k^{i \to \text{upper}}$ and $S_k^{i \to \text{lower}}$. Now, we project all 3D points in layer $S_k^{i \to \text{lower}}$ onto the $x$-$o$-$y$ plane, and we obtain a set of corresponding 2D points on this plane, denoted by $S_k^{i \to \text{lower}}$. Then, we employ the Graham's Scan [35] to construct a convex hull $\hat{S}_k^{i \to \text{lower}}$ that encircles the 2D points $S_k^{i \to \text{lower}}$. Apparently, this convex hull $\hat{S}_k^{i \to \text{lower}}$ reflects the 2D profile of the 3D points in layer $S_k^{i \to \text{lower}}$. Similarly, we achieve convex hulls $\hat{S}_k^{i \to \text{upper}}$, $\hat{S}_{k-1}^{i \to \text{lower}}$ and $\hat{S}_{k-1}^{i \to \text{upper}}$, for the 3D points in layers $S_k^{i \to \text{upper}}$, $S_{k-1}^{i \to \text{lower}}$, and $S_{k-1}^{i \to \text{upper}}$, respectively. Fig. 2 illustrates one matched pair of clusters, representing the $i$-th landmark, in two consecutive frames $k$–1 and $k$. In the left subfigure, the deep green points denote the upper layer of the cluster in frame $k$–1 (i.e. $S_{k-1}^{i \to \text{upper}}$), with its convex hull $\hat{S}_{k-1}^{i \to \text{upper}}$ outlined in red; The light green points represent the lower layer of the cluster in frame $k$–1 (i.e. $S_{k-1}^{i \to \text{lower}}$), and its convex hull $\hat{S}_{k-1}^{i \to \text{lower}}$ is delineated in blue. Similarly, in the right subfigure, the deep green points denote the upper layer of the cluster in frame $k$ (i.e. $S_k^{i \to \text{upper}}$), with its convex hull $\hat{S}_k^{i \to \text{upper}}$ outlined in red; The yellow points represent the lower layer of the cluster in frame $k$ (i.e. $S_k^{i \to \text{lower}}$), and its convex hull $\hat{S}_k^{i \to \text{lower}}$ is delineated in blue.

*c) Evaluation of Convex Hull Similarity*

In general, landmarks with high robustness (i.e. landmarks that can be consistently and stably detected in consecutive frames) significantly facilitate localization of autonomous vehicles/robots based on landmark registration. The core of landmark registration algorithms lies in the evaluation of landmark similarities, which indeed reflects the robustness of landmarks. This paper proposes a comprehensive index to evaluate the similarity of convex hulls based on the turning function [36] and the Hausdorff distance [37]. This index plays a key role in our proposed L-SLAM approach and its details are elaborated as follows.

We now discuss the turning function for a certain type of polygons – convex hulls. A turning function can be used to characterize the orientations of the sides of a convex hull. Specifically, a turning function measures the angle between each side of a convex hull and some reference orientation associated with this convex hull [36]. By this means, the turning function records the 'turning' of each side, with its value increasing for counterclockwise turns and decreasing for clockwise turns.

The detailed explanation on the turning function of a polygon can be found in [36]. As for a convex hull, the argument of its turning function is the normalized side length (i.e. the total perimeter length is scaled to 1) measured from some reference point on this convex hull, and the range of the argument is also scaled to [0, 1]. The value of this turning function, ranging within [0, 2π], is the cumulative sum of the angles that the sides of a convex hull make with some reference orientation, and these angles are usually termed as exterior angles [38]. Note that for a convex hull, its turning function is always monotonically increasing.

Based on the above definitions, in this study, the area formed between two turning function curves is used as a measure of the similarity between two convex hulls.

The turning function of a convex hull is obtained as follows. Firstly, the side lengths of this convex hull are calculated and normalized as described above. Secondly, the vertex of the convex hull nearest to the origin (i.e. LiDAR center) is chosen as the reference point for measuring side length, and the exterior angle associated with this vertex (i.e. the exterior angle formed by the two sides which make this vertex), as shown in Fig. 3., is recorded as the initial value of the turning function. Then, the exterior angles associated with other vertexes are found counterclockwise and summed up, until this sum reaches 2π. Upon completing the above three steps, we achieve the turning function of a convex hull with its argument being the normalized side length and the function value being the cumulative exterior angle.

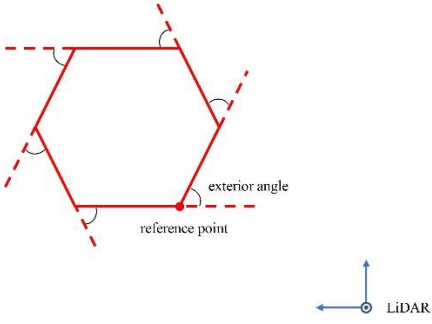

**Fig. 3**. Examples of exterior angles convex hull.

Fig. 4 demonstrates two example convex hulls constructed in two consecutive frames $k$–1 and $k$, and Fig. 5 shows the obtained turning functions of these two convex hulls based on the above approach. As previously mentioned, the turning function records the 'turning' of each side of a convex hull, which apparently reflects the characteristics of the convex hull's shape. Hence, we might as well define the following index to quantify the 'shape similarity' of two convex hulls:

$$\delta(S_{k-1}^{i \to per}, S_k^{i \to per}) = \int_0^1 \left[ f_{S_{k-1}^{i \to per}}(x) - h_{S_k^{i \to per}}(x) \right] dx \quad (5)$$

where *per* is an index that indicates the layer of a convex hull (upper layer or lower layer) with $per \in \{upper, lower\}$, and $f_{S_{k-1}^{i \to per}}(x)$ denotes the turning function of the $i$-th pair of clusters in frame $k$–1, $h_{S_k^{i \to per}}(x)$ denotes the turning function of the $i$-th pair of clusters in frame $k$, and $\delta(S_{k-1}^{i \to per}, S_k^{i \to per})$ denotes the difference between the areas formed by the two turning function curves and the horizontal axis. Apparently, a smaller difference indicates a greater similarity between convex hulls $S_{k-1}^{i \to per}$ and $S_k^{i \to per}$.

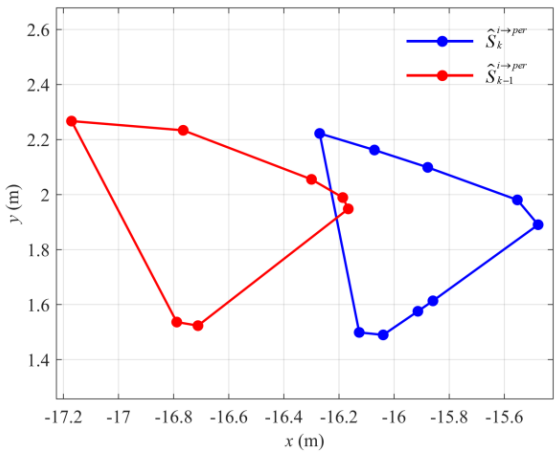

**Fig. 4**. Illustration of Convex Hulls in Two Consecutive Frames.

As seen above, in this study the turning functions serve as a measure of convex hull similarity in terms of geometrical congruence. It should be noted that turning functions are only dependent on the shapes of convex hulls, and these functions remain invariant under convex hull scaling. Namely, convex hulls with identical shapes but distinct sizes can lead to exactly the same turning function curves. However, in SLAM applications, the robustness of landmarks should be evaluated in terms of both shape similarity and dimensional similarity. Hence, apart from turning function, the following Hausdorff distance is employed in this study to measure landmark similarity in terms of convex hull dimensions.

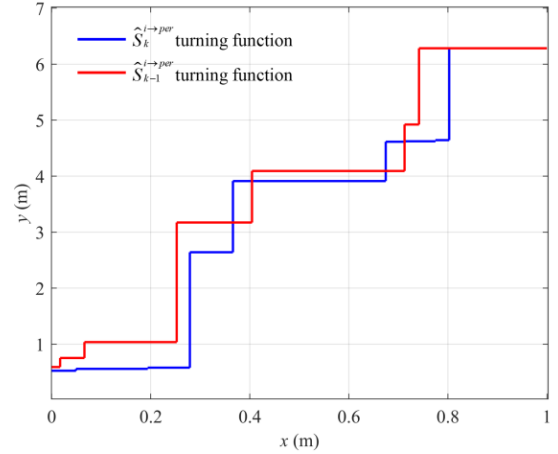

**Fig. 5**. Turning Functions of Convex Hulls.

In this study, the Hausdorff distance [37] is used as a measure to evaluate the dimensional similarity of convex hulls. The detailed evaluation process is explained as follows. Firstly, for any vertex $a_m \in S_{k-1}^{i \to per}$, the Euclidean distance between $a_m$ and each vertex on convex hull $S_k^{i \to per}$ is computed and the smallest distance, i.e. $\min_{b_n \in S_k^{i \to per}} |a_m - b_n|$, is obtained and recorded.

Repeating the above process for the $p$ vertices of convex hull $S_{k-1}^{i \to per}$, we obtain $p$ minimum distances accordingly. The maximum of these $p$ minimum distances is the unidirectional Hausdorff distance from $S_{k-1}^{i \to per}$ to $S_k^{i \to per}$, which can be expressed as follows:

$$d(S_{k-1}^{i \to per}, S_k^{i \to per}) = \max_{a_m \in S_{k-1}^{i \to per}} \min_{b_n \in S_k^{i \to per}} \|a_m - b_n\| \quad (6)$$

where $\|\cdot\|$ denotes the Euclidean distance between two convex hull vertices. Similarly, the unidirectional Hausdorff distance from $S_k^{i \to per}$ to $S_{k-1}^{i \to per}$ is given by:

$$d(S_k^{i \to per}, S_{k-1}^{i \to per}) = \max_{b_n \in S_k^{i \to per}} \min_{a_m \in S_{k-1}^{i \to per}} \|b_n - a_m\| \quad (7)$$

Then, the larger one of $d(S_{k-1}^{i \to per}, S_k^{i \to per})$ and $d(S_k^{i \to per}, S_{k-1}^{i \to per})$ is chosen as the final Hausdorff distance $H(S_{k-1}^{i \to per}, S_k^{i \to per})$, namely:

$$H(S_{k-1}^{i \to per}, S_k^{i \to per}) = \max[h(S_{k-1}^{i \to per}, S_k^{i \to per}), h(S_k^{i \to per}, S_{k-1}^{i \to per})] \quad (8)$$

The Hausdorff distance measures the level of dimensional similarity between two convex hulls. Namely, a smaller value of Hausdorff distance indicates less dimensional disparity between two convex hulls.

Based on the above tools – turning function and Hausdorff distance, in this section we propose a comprehensive similarity index to evaluate the similarity of convex hulls, taking into account both shape and dimension of landmarks. This comprehensive similarity index is mathematically given by:

$$Sim(S_{k-1}^{i \to per}, S_k^{i \to per}) = \alpha \times \delta(S_{k-1}^{i \to per}, S_k^{i \to per}) + \beta \times H(S_{k-1}^{i \to per}, S_k^{i \to per}) \quad (9)$$

where $\alpha$ and $\beta$ ($\alpha + \beta = 1$) denote the weighting factors for shape similarity and dimensional similarity, respectively. In this study, these two types of similarities are equally weighted, namely $\alpha = \beta = 0.5$.

The smaller this comprehensive similarity index is, the more similar the convex hulls are. Based on this index, we can obtain convex hull pairs with maximum similarity, $S_{k-1}^{i \to sel}$ and $S_k^{i \to sel}$, where $sel \in \{upper, lower\}$.

*d) Pose Estimation*

In this section, we explain in detail how pose estimation is achieved by maximizing the extent of overlap between similar convex hulls constructed in two consecutive frames, by means of the coordinate rotation transformation algorithm.

Assuming two convex hulls $A$ and $B$, it can be proven that their intersection $A \cap B$ is still a convex hull. A vertex of convex hull $A \cap B$ can have two possible situations of locations: (1) inside convex hull $A$ or $B$, or (2) on the edge of convex hull $A$ or $B$ (i.e. the intersecting point). These two situations are demonstrated in Fig. 6.

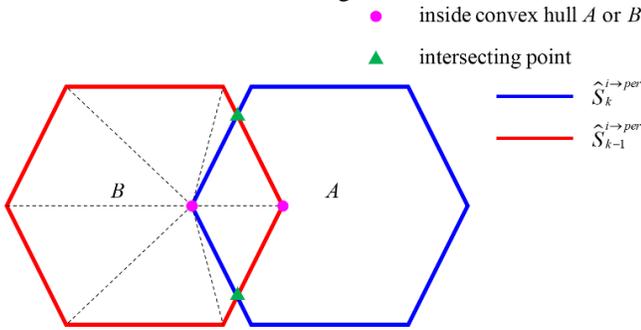

Fig. 6. Intersection and Vertex Identification of Convex Hulls

We are interested to find out all vertices that belong to convex hull $A \cap B$. Firstly, we might as well connect a vertex of $A$ with all vertices of $B$, and then we obtain a number of triangles as shown in Fig. 7. If the sum of all triangle areas is greater than the area of $B$, then this vertex of $A$ is located outside $B$; otherwise, it is located inside $B$ (i.e. this vertex also belongs to $A \cap B$). Following this approach, the vertices of $A \cap B$ that are inside convex hull $A$ or $B$ (i.e. situation (1)) can be identified. Secondly, the remaining vertices of $A \cap B$, which are located on convex hull edges (i.e. situation (2)), can be readily obtained by finding out all intersecting edges of $A$ and $B$. Upon obtaining all vertices of $A \cap B$, they are sorted in the counterclockwise direction, and a convex hull $(A \cap B)_{ordered}$ with ordered vertices is then achieved.

The area of a convex hull with ordered vertices can be computed using the following function:

$$\text{Area}(C) = \frac{1}{2} \sum_{r=1}^{M} \left( x_r y_{(r+1)} - x_{(r+1)} y_r \right) \quad (10)$$

where $C$ represents a convex hull with ordered vertices, $M$ denotes the number of convex hull vertices, $(x_r, y_r)$ ($r = 1, 2, ..., M$) stands for the coordinates of a vertex. Note that we have $x_{(M+1)} = x_1$ and $y_{(M+1)} = y_1$.

In this paragraph, we investigate how a cost function can be designed to achieve pose estimation, based on the overlapping area discussed above. We define the amount of pose change as $\boldsymbol{T} = [\Delta x, \Delta y, \Delta \theta]^T \in \mathbb{R}^{3 \times 1}$, where $\Delta x$ and $\Delta y$ denote the amount of position change in the $x$ and $y$ directions respectively, and $\Delta \theta$ denotes the amount of change in the heading angle. In our odometry, optimal vehicle pose is estimated by maximizing the overlapping area of similar convex hulls constructed in two consecutive frames. Specifically, the similarity of convex hull $S_{k-1}^{i \to sel}$ in frame $k-1$ and its associated convex hull $S_k^{i \to sel}$ in frame $k$ should be evaluated in a uniform coordinate system. Hence, $S_k^{i \to sel}$ is first transformed to the vehicle coordinate system in frame $k-1$, and then compared with $S_{k-1}^{i \to sel}$ in terms of overlapping area. This process can be mathematically expressed as follows:

$$g_i(\boldsymbol{T}) = \text{Area}\left( \left( S_k^{i \to sel} \otimes \boldsymbol{T} \right) \cap S_{k-1}^{i \to sel} \right) \quad (11)$$

where $\otimes$ is a coordinate transformation operator, and $g_i(\boldsymbol{T})$ denotes the overlapping area of the $i$-th pair of matched convex hulls after coordinate transformation. Note that this operation, $S_k^{i \to sel} \otimes \boldsymbol{T}$, transforms convex hull $S_k^{i \to sel}$ from the vehicle coordinate system in frame $k$ to that in frame $k-1$, using the vehicle pose change information $\boldsymbol{T}$. Now, the following cost function can be constructed for pose estimation purposes:

$$G(\boldsymbol{T}) = \underset{\boldsymbol{T}}{\text{argmax}} \left( \sum_{i=1}^{N} g_i(\boldsymbol{T}) \right) \quad (12)$$

where $N$ is the total number of matched convex hull pairs in two consecutive frames. The objective of this cost function is to find an optimal vehicle pose change $\boldsymbol{T}$ such that the sum of all overlapping areas of the $N$ convex hull pairs is maximized.

The step size of iteration plays an important role in the odometry design, a suitable iteration step size reduces the amount of computation and accelerates the optimization process.

In this study, the step size of iteration is determined according to the following equation:

$$\alpha = C \times \text{ceil}\left(100 \times \left(1 - \frac{\sum_{i=1}^{N} g_i(\mathbf{T})}{\sum_{i=1}^{N} p_i}\right)\right) \quad (13)$$

where $\alpha$ denotes the step size, $C$ is a user-defined coefficient whose value can be different for translation and rotation, $\text{ceil}(\cdot)$ is a rounding function which computes the nearest integer greater than (or equal to) its argument, $p_i$ represents the smaller (in area) convex hull of $S_{k-1}^{i \to \text{sel}}$ and $S_k^{i \to \text{sel}} \otimes \mathbf{T}$. In the initial phase of iteration, as the pose change $\mathbf{T}$ is far from the optimal solution, the corresponding function value $G(\mathbf{T})$ is relatively large, which ultimately leads to a larger iteration step size $\alpha$. With the increase in the number of iterations, the pose change $\mathbf{T}$ approaches the optimal solution, and the corresponding objective function value $G(\mathbf{T})$ increases. Eventually, the iteration step size $\alpha$ becomes smaller until the iteration converges to the optimal solution. The adoption of such an adaptive iteration step size can accelerate the process of the algorithm to obtain the optimal solution and reduce the time consumed in iteration.

As mentioned above, a cost function has been established to achieve optimal vehicle pose by maximizing the overall overlapping area of the $N$ convex hull pairs. This cost function can be solved by means of the coordinate rotation method [11]. Specifically, only one variable in $\mathbf{T} = [\Delta x, \Delta y, \Delta \theta]^T$ (i.e. $\Delta x$, $\Delta y$, or $\Delta \theta$) is changed at a time, and the other variables are kept constant. The search for the maximum value of the cost function $G(\mathbf{T})$ is performed in the $\Delta x - \Delta y - \Delta \theta$ coordinate system, along the direction of the three orthogonal axes in turn.

For example, we define the search direction as $\mathbf{d}_u^v$ if the search is performed in the $v$-th round along the $u$-th coordinate axis. Accordingly, this iteration can be mathematically expressed as:

$$\mathbf{T}_u^v = \mathbf{T}_{u-1}^v + \alpha \mathbf{d}_u^v \quad (v = 0, 1, 2, ..., n; u = 1, 2, 3) \quad (14)$$

Note that in this equation, the $\Delta x$-axis is the 1st coordinate axis, the $\Delta y$-axis is the 2nd coordinate axis, and the $\Delta \theta$-axis is the 3rd coordinate axis. Besides, to ensure effective coordinate rotation, we also define that $\mathbf{T}_0^v = \mathbf{T}_3^{v-1}$. In the iteration process, if $\left\| \mathbf{T}_u^v - \mathbf{T}_0^v \right\| < \varepsilon$ ($\varepsilon$ denotes the threshold for terminating iteration) then we perform $\mathbf{T}^* \leftarrow \mathbf{T}_n^v$; Otherwise we have $\mathbf{T}_0^{v+1} \leftarrow \mathbf{T}_n^v$, and the next round of search is performed until the termination condition is met. By this means, the above cost function (12) can be solved, which in turn leads to the optimal pose estimation of the vehicle.

*2) Vertical Pose Estimation*

In this section, we expound upon the principles of estimating the vehicle pose in the vertical plane, i.e. position in the $z$-direction and pitch angle with respect to its initial pose. The pitch angle is estimated using point clouds reflected by the ground in front of and behind the vehicle. By executing plane fitting and calculating the normal vectors of these fitted planes, the vehicle's pitch angle with respect to its initial pose can be computed. Then, the vehicle position in the $z$-direction can be determined, given the estimated pitch angle, yaw angle and horizontal positions.

*a) Pitch Angle Estimation*

In the course of navigation, changes in the pitch angle of a vehicle correspond to simultaneous variations in the normal vectors of ground planes. Hence, the vehicle pitch angle can be inferred from the normal vectors of the fitted planes in front of and behind the vehicle. As shown in Fig. 7, the normal vectors $n_f$ and $n_r$ of the front fitted plane (in blue color) and the rear fitted plane (in red color) form an included angle $\Delta \psi$, which value is treated as the amount of pitch angle change. Thus, the overall vehicle pitch angle with respect to its initial pose can be determined by summing up all historical pitch angle changes. The above-mentioned pitch angle change, $\Delta \psi$, can be mathematically expressed as follows:

$$\Delta \psi = \arccos\left(\frac{n_f \cdot n_r}{\|n_f\| \cdot \|n_r\|}\right) \quad (15)$$

where the operator "$\cdot$" denotes the dot product of two vectors, and the operator $\|\cdot\|$ represents the norm, or the length, of a vector.

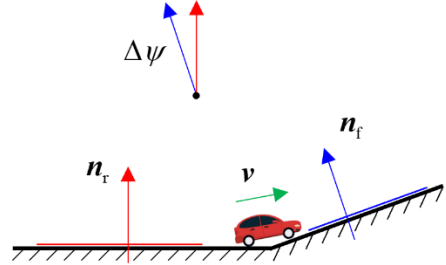

**Fig. 7**. Normal vectors (and their included angle) of planes fitted to the forward-facing and rearward-facing point clouds of the vehicle.

*b) Vertical Position Estimation*

The position of a vehicle in the $z$-direction, with respect to its initial pose, is dependent upon the previously estimated pitch angle, yaw angle and horizontal positions. Specifically, the vertical position change of a vehicle at time $k$ with respect to time $k$–1, expressed by $\Delta z_k$, can be written as follows:

$$\Delta z_k = \left|\sin(\sum_{i=1}^{k} \Delta \psi_k)\right| \cos(\sum_{i=1}^{k} \Delta \theta_i) \sqrt{\Delta x_k^2 + \Delta y_k^2} \quad (16)$$

where $\Delta x_k$ and $\Delta y_k$ denote the amount of position change in the $x$ and $y$ directions at time $k$, $\Delta \psi_k$ represents the pitch angle change at time $k$, and $\Delta \theta_k$ indicates the heading angle change at time $k$. Similar to pitch angle, the overall vehicle vertical position with respect to its initial pose can be determined by summing up all historical vertical position changes.

*C. Mapping*

The primary objective of mapping is to present a comprehensive depiction of the surrounding environment, thereby providing a constructed global map for path planning and navigation.

In the SLAM process, the vehicle/robot constantly receives point cloud measurements in each frame, and these points are obtained with respect to a moving coordinate system, i.e. the vehicle coordinate system. To complete a global map, we must transform points obtained in each frame to a consistent world coordinate system by means of transform integration. This process aligns the point cloud (both the ground and the landmarks) collected in each frame with the incremental global map. Following this step, the global map undergoes a voxelization process, thereby achieving a lightweight global map with rich details in the form of voxel grids.

Our proposed method is able to combine point clouds of both landmarks and ground to construct a global map, thereby providing the resulting global map with rich details. Specifically, compared to commonly used map types – geometric feature maps (e.g., maps with plane features [39], maps with line segments [22] and maps with points [40]) and occupancy grid maps [33], our method better preserves the appearance information of landmarks by performing landmark layering and establishing convex hulls that reflect the exterior profiles of landmarks. Note that the process of landmark layering and convex hull establishment introduce more details of the environment in the constructed map, yielding a more comprehensive and sophisticated representation of the explored environment.

## IV. EVALUATION OF EXPERIMENTAL RESULTS

In this study, we utilized two different types of point cloud sources, the KITTI dataset [13] and point clouds collected by our Unmanned Ground Vehicle (UGV) platform, to evaluate the effectiveness of our proposed LiDAR odometry. In terms of the KITTI dataset, our assessment was primarily based on two official metrics, the average translational error ($t_{rel}$) and the average rotational error ($r_{rel}$), to demonstrate the superiority of our proposed L-LO. Several typical odometry methods were employed for comparison purposes, using the same KITTI sequences. As for tests with the UGV, the typical LiDAR odometry LOAM-Velodyne [12] was implemented and used as a bench mark for field evaluation. Quantitative metrics including the average error (AE) of the estimated position, the standard deviation (SD) of the estimated position, and the position error of the trajectory endpoints (PETE), were introduced to evaluate these two competing methods. We conducted our comparative studies on a desktop with an Intel Xeon Gold 6128 CPU, a 64 GB RAM and an Nvidia Quadro P5000 GPU with 16 GB RAM.

*A. Tests with KITTI Datasets*

We evaluated our proposed method, L-LO, against several typical techniques using the KITTI odometry sequences 00, 02–10. These point cloud data were obtained by a Velodyne HDL-64ES2 LiDAR, operating at a frequency of 10 Hz and 360 horizontal FOV. Various types of environments are covered in these sequences, including residential, urban, and campus.

The results summarized in Table I provide a comparative analysis of our proposed method with other existing typical methods such as LeGO-LOAM [17], SuMa F-to-F [41], MULLS-LO(s1) [42], ICP-Point2Point [43], ICP-point2plane [44], A-LOAM w/o mapping, G-ICP [45]. Velas et al. [46] and LOAM-Velodyne [12]. Based on the results presented in Table I, we see that the proposed L-LO method outperforms most of the competing methods in terms of localization accuracy. In particular, the L-LO achieves the highest average localization accuracy among all methods compared.

Fig. 8 demonstrates visual comparison results between the ground truths and the estimated trajectories produced by the proposed L-LO algorithm. The blue curves correspond to the ground truth trajectories, while the red curves represent the estimated trajectories computed by our L-LO algorithm. It is

TABLE I
LOCALIZATION ACCURACY COMPARISON BETWEEN THE PROPOSED L-LO AND EXISTING METHODS

| Seq.<br>Method | 00 | 01 | 02 | 03 | 04 | 05 | 06 | 07 | 08 | 09 | 10 | Average |
|---|---|---|---|---|---|---|---|---|---|---|---|---|
| | $t_{rel}/r_{rel}$ | $t_{rel}/r_{rel}$ | $t_{rel}/r_{rel}$ | $t_{rel}/r_{rel}$ | $t_{rel}/r_{rel}$ | $t_{rel}/r_{rel}$ | $t_{rel}/r_{rel}$ | $t_{rel}/r_{rel}$ | $t_{rel}/r_{rel}$ | $t_{rel}/r_{rel}$ | $t_{rel}/r_{rel}$ | $t_{rel}/r_{rel}$ |
| LeGO-LOAM | 2.17/1.05 | 13.4/1.02 | **2.17**/1.01 | 2.34/1.18 | 1.27/1.01 | 1.28/0.74 | 1.06/0.63 | 1.12/0.81 | 1.99/0.94 | 1.97/0.98 | 2.21/0.92 | 2.49/1.00 |
| SuMa F-to-F | 2.10/0.90 | 4.00/1.20 | 2.30/0.80 | **1.40**/0.70 | 11.90/1.10 | 1.50/0.80 | 1.00/0.60 | 1.80/1.20 | 2.50/1.00 | 1.90/0.80 | **1.80**/1.00 | 2.93/0.92 |
| MULLS-LO(*s1*) | 2.36/1.06 | 2.76/0.89 | 2.81/0.95 | 1.26/0.67 | 5.72/1.15 | 2.19/1.01 | 1.12/0.51 | 1.65/1.27 | 2.73/1.19 | 2.14/0.96 | 3.61/1.65 | 2.57/1.03 |
| ICP-Point2Point | 6.88/2.99 | 11.21/2.58 | 8.21/3.39 | 11.07/5.05 | 6.64/4.02 | 3.97/1.93 | 1.95/1.59 | 5.17/3.35 | 10.04/4.93 | 6.93/2.89 | 8.91/4.74 | 7.37/3.38 |
| ICP-point2plane | 3.80/1.73 | 13.53/2.58 | 9.00/2.74 | 2.72/1.63 | 2.96/2.58 | 2.29/1.08 | 1.77/1.00 | 1.55/1.42 | 4.42/2.14 | 3.95/1.71 | 6.13/2.60 | 4.74/1.93 |
| CLS | 2.11/0.95 | 4.22/1.05 | 2.29/0.86 | 1.63/1.09 | 1.59/0.71 | 1.98/0.92 | 0.92/0.46 | 1.04/0.73 | 2.14/1.05 | 1.95/0.92 | 3.46/1.28 | 2.12/0.91 |
| A-LOAM w/o mapping | 4.08/1.69 | 3.31/0.92 | 7.33/2.51 | 4.31/2.11 | 1.60/1.13 | 4.09/1.68 | 1.03/0.52 | 2.89/1.8 | 4.82/2.08 | 5.76/1.85 | 3.61/1.76 | 3.89/1.64 |
| G-ICP | 1.29/0.64 | 4.39/0.91 | 2.53/0.77 | 1.68/1.08 | 3.76/1.07 | 1.02/0.54 | 0.92/0.46 | 0.64/0.45 | 1.58/0.75 | 1.97/**0.77** | 1.31/**0.62** | 1.92/0.73 |
| Velas et al. | 3.02/- | 4.44/- | 3.42/- | 4.94/- | 1.77/- | 2.35/- | 1.88/- | 1.77/- | 2.89/- | 4.94/- | 3.27/- | 3.15/- |
| LOAM Velodyne | 3.41/- | 6.54/- | 5.66/- | 1.64/- | **1.09**/- | 1.32/- | 1.01/- | 1.26/- | 2.16/- | 1.44/- | 1.91/- | 2.49/- |
| L-LO | **0.91**/**0.81** | NA/NA | 2.35/0.87 | 2.92/**0.34** | 2.28/**0.42** | **0.82**/**0.31** | **0.66**/**0.37** | **0.69**/**0.26** | **1.37**/**0.58** | **1.89**/0.94 | 1.90/0.75 | **1.59**/**0.57** |

<span style="color:red">Note: on the official website of KITTI dataset, two important metrics – average translational error $t_{rel}$ (%) and average rotational error $r_{rel}$ (°/100 m) – are introduced to evaluate the localization performance of different odometry solutions. Following this convention, these two metrics are also employed in this study for odometry performance evaluation. Bold numbers indicate the best performance in the corresponding column of this table.</span>

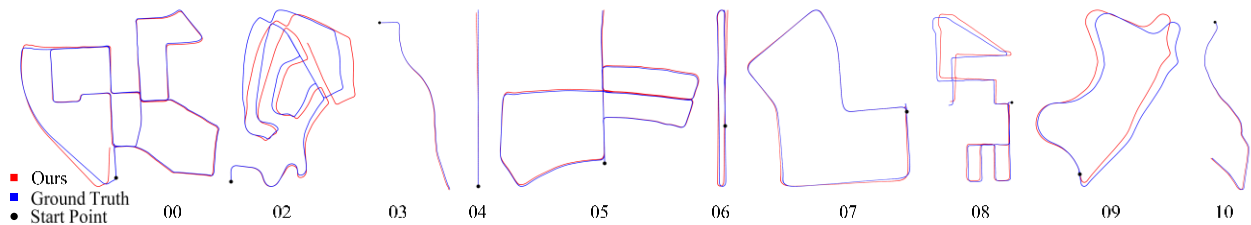

**Fig. 8**. Trajectories of the proposed L-LO and the ground truths on KITTI sequences 00, 02-10.

seen in Fig. 8 that for most sequences close alignment between the estimated and ground truth trajectories can be achieved. Specifically, in residential areas (sequences 00, 05, 06, 07, 08, 09), our proposed L-LO algorithm exhibits impressive capacity in accurately estimating the vehicle trajectories with minimal deviations from the ground truths. As an illustration of the mapping performance, the KITTI sequence 07 was utilized to establish a global map with distinct colors indicating different landmarks (see Fig. 9).

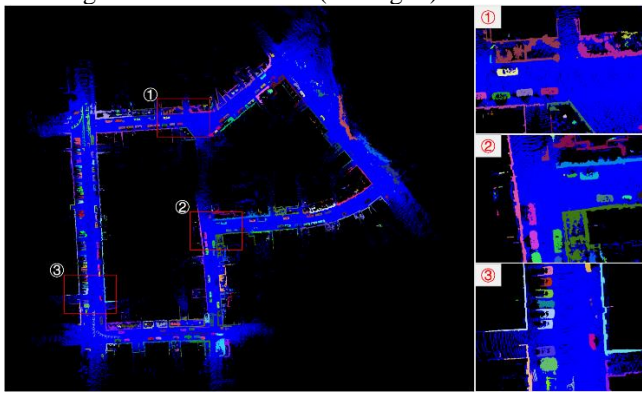

**Fig. 9.** Using only LiDAR scans, our approach generated a map of sequence 07 in the KITTI dataset. This map is represented by 3D grids that are color-coded to indicate different landmarks. Our odometry provides high-quality maps with superior metric accuracy and landmark identification compared to its non-color-coded counterparts.

*B. Tests with UGV*

Apart from experiments using public datasets, we also evaluated our L-LO using data collected by a UGV platform equipped with a RoboSense 16-beam LiDAR and a CHC CGI-630 GNSS. This UGV platform is shown in Fig. 10. The RoboSense 16-beam LiDAR was used to collect point cloud data for the proposed odometry and its competitor, while the CHC CGI-630 GNSS was employed to provide the ground truth trajectories of the UGV platform. Extensive tests were conducted in four different scenes within the Chongqing University campus at a low velocity using the UGV platform. These four campus scenes are denoted by CQU 01, CQU 02, CQU 03 and CQU 04, and the lengths of the ground truth trajectories in these four scenes are 162.41 m, 289.98 m, 542.43 m, and 177.67 m, respectively.

*1) Localization Accuracy Assessment*

Using the data collected within Chongqing University, the performance of the proposed L-LO was evaluated in comparison with the LOAM-Velodyne method and the ground truth. Fig. 11 shows the UGV trajectories traversed in four different campus scenes, where the black curves represent the ground truth trajectories, the red curves denote the estimated trajectories of the L-LO, and the blue curves stand for the estimated trajectories of the LOAM-Velodyne. It can be clearly seen from this figure that the proposed L-LO consistently outperforms the competing LOAM-Velodyne for all four campus scenes.

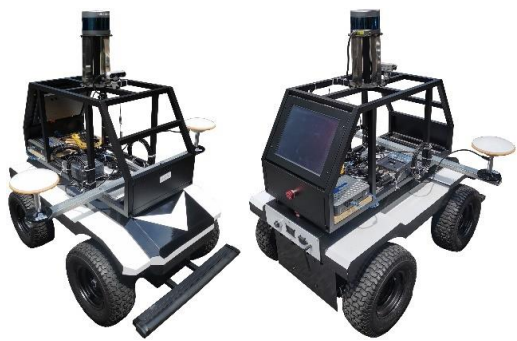

**Fig. 10**. UGV experimental platform equipped with a RoboSense 16-beam LiDAR and a CHC CGI-630 GNSS.

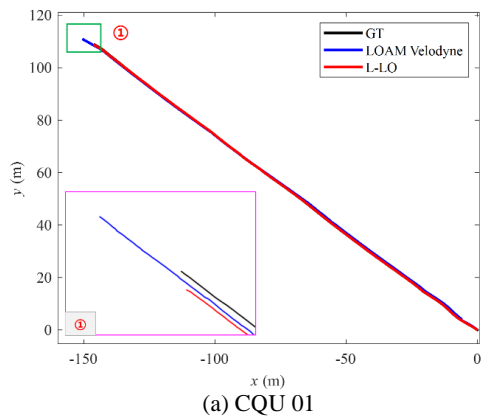

(a) CQU 01

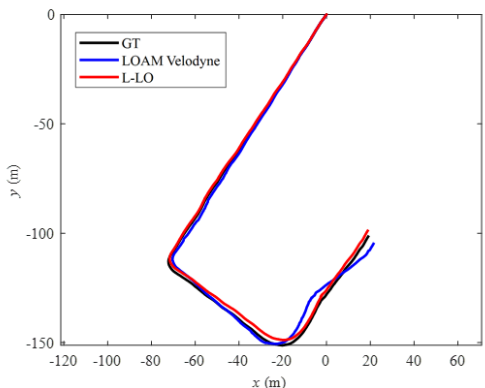

(b) CQU 02

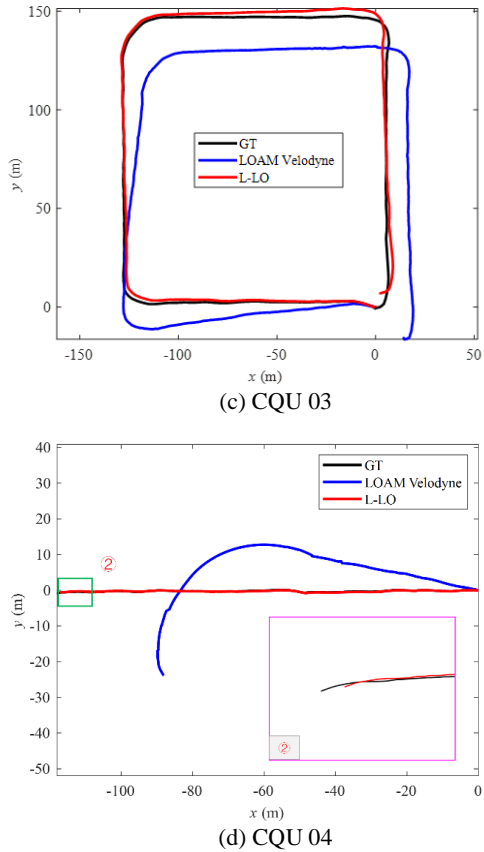

(c) CQU 03

(d) CQU 04

**Fig. 11.** UGV trajectory estimation results using the test data of four different scenes within the campus.

TABLE II
LOCALIZATION ACCURACY COMPARISON BETWEEN THE PROPOSED L-LO AND THE LOAM-VELODYNE

| Seq. | | CQU 01 162.41m | CQU 02 289.98m | CQU 03 542.43m | CQU 04 177.67m | Ave. |
|---|---|---|---|---|---|---|
| LOAM-Velodyne | AE | 4.30 | 9.75 | 17.92 | 10.75 | 10.68 |
| | SD | 2.12 | 4.47 | 7.19 | 8.21 | 5.50 |
| | PETE | 9.16 | 7.33 | 24.99 | 38.28 | 19.94 |
| L-LO | AE | 1.35 | 1.55 | 3.39 | 0.32 | 1.65 |
| | SD | 0.52 | 0.86 | 2.55 | 0.16 | 1.02 |
| | PETE | 0.98 | 2.67 | 8.18 | 0.71 | 3.14 |

Table II demonstrates the quantitative results of the proposed L-LO and the LOAM-Velodyne. Several important metrics that characterize the localization accuracy are shown in this table, including the average error (AE) of the estimated position, the standard deviation (SD) of the estimated position, and the position error of the trajectory endpoints (PETE). We see in this table that for all four campus scenes, the proposed L-LO consistently outperforms its competitor by providing substantially smaller AE, SD and PETE.

*2) Runtime Evaluation*

To evaluate the time consumption of the proposed algorithm, we evaluated individually the execution time of the three main modules: "Pre-processing", "LiDAR Odometry", and "Mapping" on four data sequences: CQU 01, CQU 02, CQU 03, and CQU 04, as detailed in Table III.

In this table, columns labeled "CQU 01", "CQU 02", "CQU 03", and "CQU 04" denote the average runtime (in ms) per-frame of our proposed algorithm across each respective data sequence. For each data sequence, the runtime breaks down into three primary modules: "Pre-processing", "LiDAR Odometry", and "Mapping". The last row – Total Runtime – sums up the average runtime of these three modules for each data sequence. The last column of the table – Average – offers an average runtime (in ms) per-frame of our proposed algorithm across all four data sequences.

TABLE III
RUNTIME EVALUATION ON DIFFERENT MODULES OF THE PROPOSED METHOD

| Data Seq. Module | CQU 01 | CQU 02 | CQU 03 | CQU 04 | Ave. |
|---|---|---|---|---|---|
| Per-processing | 137.44 | 153.30 | 107.59 | 117.79 | 129.03 |
| LiDAR Odometry | 45.55 | 55.52 | 42.79 | 50.25 | 48.53 |
| Mapping | 43.10 | 53.71 | 35.12 | 34.17 | 41.53 |
| Total Time | 226.09 | 262.53 | 185.50 | 202.21 | 219.09 |

The data in Table III clearly reveals that the "Pre-processing" module accounts for the dominant share of the overall runtime. As seen in the last column, the average runtime of this module across all sequences is approximately 50% longer than that of the other two modules combined. In comparison, the key contribution of this paper – LiDAR Odometry – exhibits considerably faster execution speed, achieving an average runtime of 48.57 ms only (over 20 Hz) across all four data sequences. Additionally, the "Mapping" module requires the least runtime among the three modules in the proposed algorithm. In our next step of research, optimization endeavors will be concentrated on the most time-consuming module – Pre-processing, aiming to enhance the overall real-time performance.

*C. Ablation Experiment*

The ablation experiments in this section aim to validate the efficacy of two key techniques in our proposed approach: 1) whether the accuracy of pose estimation can be improved by using the comprehensive similarity index to find the more similar pair of convex hulls for registration, and 2) whether adopting an adaptive iteration step size for cost function optimization can enhance computational efficiency.

*1) Advantage of Comprehensive Similarity Index*

In this ablation experiment, we performed pose estimation using three types of convex hulls for registration: 1) convex hull pairs determined by the comprehensive similarity index, 2) convex hull pairs only in the upper-layer, and 3) convex hull pairs only in the lower-layer. These three comparative tests were conducted on four data sequences: CQU 01, CQU 02, CQU 03, and CQU 04. Also, the localization accuracy was evaluated using three metrics, i.e. AE, SD, and PETE.

Table IV shows the results of these three comparative tests. It is observed that using convex hull pairs determined by the comprehensive similarity index achieves lower AE, SD, and PETE, compared to the results obtained by using only the upper-layer or lower-layer convex hull pairs for registration. This indicates that by using the proposed index, we are able to find the pair of convex hulls with higher similarity for registration purpose, which leads to reduced registration error and higher localization accuracy.

TABLE IV
LOCALIZATION ACCURACY USING DIFFERENT CONVEX HULL PAIRS FOR REGISTRATION

| Data Seq. / Choice of CHPs | | CQU 01 | CQU 02 | CQU 03 | CQU 04 | Ave. |
|---|---|---|---|---|---|---|
| CHPs determined by the comprehensive similarity index | AE | 1.35 | 1.55 | 3.39 | 0.32 | 1.65 |
| | SD | 0.52 | 0.86 | 2.55 | 0.16 | 1.02 |
| | PETE | 0.98 | 2.67 | 8.18 | 0.71 | 3.14 |
| CHPs in the lower-layer only | AE | 1.72 | 1.67 | 4.52 | 0.48 | 2.10 |
| | SD | 0.65 | 1.88 | 3.81 | 0.81 | 1.78 |
| | PETE | 1.78 | 4.63 | 11.54 | 1.59 | 4.89 |
| CHPs in the upper-layer only | AE | 2.01 | 1.98 | 5.60 | 0.73 | 2.58 |
| | SD | 1.08 | 1.90 | 4.31 | 1.35 | 2.16 |
| | PETE | 1.94 | 5.82 | 16.79 | 2.57 | 6.78 |

Note: In this table, CHP is short for Convex Hull Pair.

*2) Advantage of Adaptive Iteration Step Size*

In this ablation study, we conducted comparative tests using two step size configurations for cost function optimization: 1) the adaptive iteration step size proposed in this paper, and 2) a fixed iteration step size (i.e. $\alpha = 1$). These two step size choices were validated on the same data sequences, namely CQU 01, CQU 02, CQU 03, and CQU 04.

TABLE V
COMPARISON OF TIME CONSUMPTION USING TWO DIFFERENT ITERATION STEP SIZES

| Data Seq. / Iteration strategy | CQU 01 | CQU 02 | CQU 03 | CQU 04 | Ave. |
|---|---|---|---|---|---|
| Adaptive Step Size | 41.22 ms | 51.53 ms | 39.75 ms | 46.66 ms | 44.79 ms |
| Fixed Step Size | 163.59 ms | 148.72 ms | 109.21 ms | 121.36 ms | 135.72 ms |

Table V demonstrates the runtime performance of the proposed adaptive step size and the compared fixed step size. The four columns in the middle present the average runtime of each frame for a certain data sequence, while the last column demonstrates the overall average runtime across four data sequences.

As shown in this table, the adaptive iteration step size provides markedly shorter average runtime compared to the fixed step size, for all four tested data sequences.

Through the above two ablation experiments, we have proven two advantages of our proposed method: 1) the accuracy of pose estimation is improved by means of the comprehensive similarity index, and 2) adopting an adaptive iteration step size for the optimization process reduces computation time.

## V. CONCLUSION AND FUTURE WORKS

A novel landmark-based LiDAR odometry was introduced in this study. This work, to the best knowledge of the authors, is the first LiDAR odometry which evaluates the overall exterior characteristics of environmental landmarks, instead of using common geometric features such as points, lines and planes. The working principles of the proposed odometry were explained in detail, including point cloud preprocessing, horizontal pose estimation, vertical pose estimation, and mapping. In evaluating our proposed L-LO, we used the renowned KITTI dataset and the point cloud data collected by our UGV platform. The results have demonstrated superior positioning accuracy of the proposed L-LO, in comparison with typical LiDAR odometry solutions in the literature.

Although the results provided by our L-LO were promising, certain limitations should also be pointed out. In sequence 01 of the KITTI dataset, the proposed odometry failed to produce reliable localization results as this sequence represents a highway scenario where very few landmarks could be extracted from the sensor measurements. In recognition of this shortcoming, our future endeavors will be focused on introducing multi-modality information (such as images) in the algorithm and enhancing its robustness in challenging scenarios, striving to eliminate these limitations and further improve the odometry performance.

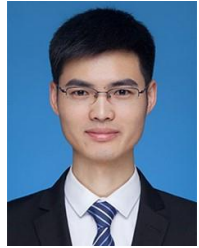

**Feiya Li** received the Bachelor's degree from Liaoning University of Technology, Liaoning, China, in 2016, and the Master's degree from Chongqing University, Chongqing, China, in 2020. He is currently working towards a Ph.D. degree at Chongqing University. His major research interests include single- and multi-vehicle SLAM solutions.

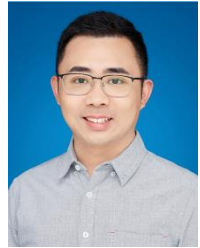

**Chunyun Fu** received the Bachelor's degree from Chongqing University, Chongqing, China, in 2010, and the Ph.D. degree from RMIT University, Victoria, Australia, in 2015. He is currently an associate professor with the College of Mechanical and Vehicle Engineering, Chongqing University. His main research interests include SLAM and multi-target tracking for autonomous vehicles, and chassis and powertrain control systems for electric vehicles.

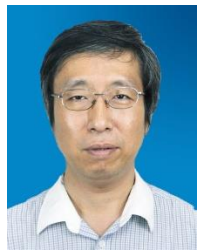

**Dongye Sun** received the Ph.D. degree from Jilin University, Changchun, China, in 1996. Since 2003, he has been a full professor with the State Key Laboratory of Mechanical Transmissions, Chongqing University, Chongqing, China. His research interests include powertrain systems and hybrid electric vehicles, gearing theory and gearing transmission systems, trajectory planning and tracking, and data mining for intelligent driving.